\title{On Text Generation Using Brain Data}
\documentclass{article}

\usepackage[preprint]{neurips_2024}

\usepackage[utf8]{inputenc} 
\usepackage[T1]{fontenc}    
\usepackage{hyperref}       
\usepackage{url}            
\usepackage{booktabs}       
\usepackage{amsfonts}       
\usepackage{nicefrac}       
\usepackage{microtype}      
\usepackage{xcolor} 
\usepackage{graphicx}
\usepackage{multirow}

\title{On Creating A Brain-To-Text Decoder}

\author{Zenon Lamprou \\
  Department of Computer Science \& Information Sciences \\
  NeuraSearch Laboratory \\
  Strathclyde University\\
  \texttt{zenon.lamprou@strath.ac.uk} \\
  \And
  Yashar Moshfeghi \\
  Department of Computer Science \& Information Sciences \\
  NeuraSearch Laboratory \\
  Strathclyde University\\
  \texttt{yashar.moshfeghi@strath.ac.uk} \\
}

\begin{document}

\maketitle

\begin{abstract}
Brain decoding has emerged as a rapidly advancing and extensively utilized technique within neuroscience. This paper centers on the application of raw electroencephalogram (EEG) signals for decoding human brain activity, offering a more expedited and efficient methodology for enhancing our understanding of the human brain. The investigation specifically scrutinizes the efficacy of brain-computer interfaces (BCI) in deciphering neural signals associated with speech production, with particular emphasis on the impact of vocabulary size, electrode density, and training data on the framework's performance. The study reveals the competitive word error rates (WERs) achievable on the Librispeech benchmark through pre-training on unlabelled data for speech processing. Furthermore, the study evaluates the efficacy of voice recognition under configurations with limited labeled data, surpassing previous state-of-the-art techniques while utilizing significantly fewer labels. Additionally, the research provides a comprehensive analysis of error patterns in voice recognition and the influence of model size and unlabelled training data. It underscores the significance of factors such as vocabulary size and electrode density in enhancing BCI performance, advocating for an increase in microelectrodes and refinement of language models.
\end{abstract}

\section{Introduction}
\label{introduction}
Deciphering the workings of the human brain has been a subject of fascination for researchers for many decades\cite{dockes_neuroquery_2020, dockes_text_2018,power_functional_2011,wager_meta-analysis_2007}. With the advent of non-invasive techniques such as EEG, MRI, and MEG, researchers have been able to study brain activations and attempt to interpret their meaning. In the early stages of research, simple binary conditions were used, such as having subjects view happy or sad scenes, to record brain activity and investigate how emotions are represented in the brain.

As research continued, more complex conditions were introduced, such as the decoding of the role of the Superior Temporal Sulcus and Human insula, as demonstrated in previous studies \cite{hein_superior_2008,chang_decoding_2013}. These studies revealed that brain areas are not responsible for a single function, but rather for a variety of functions, and that they are interconnected and work together to perform different tasks, such as audiovisual processing and language understanding.

One of the most sought-after discoveries in neuroscience is the process by which the human brain processes natural language and semantic information. Deciphering this process could provide insights into the evolution of the human brain over the last 3000 years. Neurolinguistics is a field that focuses on understanding how the human brain comprehends natural language.

Previous research in neurolinguistics \cite{toneva_interpreting_2019,wehbe_aligning_2014,reddy_can_2021, schwartz_inducing_2019} has relied primarily on gathering data using fMRI, which is slow, expensive, and not in real-time compared to other brain recording techniques. Recent studies have attempted to unravel how information needs are represented in the brain \cite{allegretti_when_2015,moshfeghi_understanding_2016}. Information needs are closely related to neurolinguistics since they are the main motivators for performing information retrieval. It is widely believed that artificial intelligence networks cannot capture the true semantic meaning and, therefore, do not provide users with the desired information.

Previous approach have been limited on classifying information need \cite{michalkova_information_2022} , mental work load \cite{kingphai_eeg_2021,kingphai_time_2023}, a reading task \cite{hollenstein_reading_2021} or an imagined category\cite{nieto_thinking_2022} to name a few . To our knowledge, this study is the first attempt to use the raw EEG signals for the purpose of identifying the exact words and generate sentences derived from the human brain. Our approach is significant because EEG is a real-time, less expensive, and easier-to-use technique compared to fMRI. By decoding the human brain using raw EEG signals, we aim to speed up the process of understanding the human brain using a faster and more efficient method. We also aim in our knowledge to be the first that utilise raw EEG signals as the input to our model to perform the decoding.

The following sections of this work are structured as follows, Section \ref{related work} provides background to the developments and importance of traditional methods for producing embeddings and their use within IR as well as how prior works have set about creating EEG-related embeddings for downstream tasks. Section \ref{methodology} discusses the methodology of the research, Section \ref{results} highlights the results of our investigation, and lastly, Section \ref{conclusion} discusses the findings of this work and their potential implications.

\section{Related work}
\label{related work}
Recent interdisciplinary research has explored the potential of neuroscience to advance information retrieval systems, conceptualized as the emerging field of NeuraSearch \cite{moshfeghi2021neurasearch}. Findings from research \cite{mueller2008electrophysiological, muller2015electroencephalography, gwizdka2019introduction, Martínez-Castaño2022} suggest that NeuraSearch seeks to enhance information accessibility and utilization by employing neurophysiological signals to deliver implicit relevance feedback, thereby augmenting user interaction in the absence of explicit inputs. Studies such as \cite{moshfeghi2013understanding} have built the foundation of NeuraSearch. For example, in \cite{moshfeghi_understanding_2016}, Moshfeghi et al utilized fMRI to explore the cerebral processes that underlie information needs during Q/A tasks, thereby mapping neural activities associated with various states of knowledge. This body of research highlights the importance of neurological insights in the customization of user experiences in information retrieval, as further validated by EEG studies like \cite{moshfeghi2019towards}, which identified EEG patterns associated with different stages of information need recognition.

The study by \cite{michalkova_information_2024} explored the enhancement of queries through the utilization of brain signals, highlighting that models driven by electroencephalography (EEG) can improve the relevance of search outcomes. This finding emphasizes the practical advantages of combining traditional query strategies with neurophysiological information to optimize search processes.

NeuraSearch prioritizes the evaluation of cognitive workload through EEG analysis, with research emphasizing its efficacy in monitoring cognitive states across various contexts \cite{Kingphai_Mental_2021}. Investigations by \cite{kingphai_eeg_2021, kingphai_time_2023} have achieved progress in the areas of emotion recognition and EEG data classification, highlighting the criticality of methodological accuracy and the necessity for standardized practices. Prominent neuroimaging modalities, including MEG and fMRI, are integral to NeuraSearch's explorations \cite{kauppi2015towards}. Their application, as illustrated in \cite{lamprou_role_2022}, explored semantic comprehension and demonstrated how neuroscience can proficiently assess cognitive processing and decision-making.

The convergence of neuroscience with information retrieval elucidates unprecedented possibilities for the development of adaptive, user-centered systems. By proficiently interpreting users' implicit feedback through neurophysiological signals, as evidenced in several pivotal studies \cite{allegretti_when_2015, jacucci2019integrating}, it facilitates the creation of more nuanced interfaces. This ongoing synthesis aims to address complex user information requirements and adapt to cognitive challenges inherent in the process of information retrieval.

Moreover recent works by \cite{kostas2021bendr} and \cite{partovi2023self} have explored the potential to produce more potent and generalized EEG representations. BENDR, introduced by \cite{kostas2021bendr}, employs transformer architectures and contrastive self-supervised learning to enhance the generation of general EEG embeddings. This approach aims to capture intricate temporal dependencies and patterns within the data. On the other hand, \cite{partovi2023self} presents a self-supervised learning framework focused on creating task-agnostic EEG embeddings, enhancing adaptability for various downstream applications.

These novel approaches offer a promising avenue to on constructing general EEG embeddings. Those embeddings are not task constrained and they learn general information about EEG signals. These approaches aim first to provide a general interface to learn EEG features without the need of using multiple pre-processing pipelines and also providing EEG features to the decoder so the decoder can translate raw EEG signals into text.

Our proposed model's architecture draws inspiration from two sources: \cite{bender2021dangers} and \cite{baevski_wav2vec_2020}. In order to achieve state-of-the-art performance in speech recognition tasks, \cite{baevski_wav2vec_2020} presents a novel framework for self-supervised learning of speech representations. Using the jointly trained wav2vec 2.0 framework, a contrastive problem is solved over quantized latent representations by masking the speech input in the latent space. With minimal labeled data, the study demonstrates the competitive word error rates (WER) that can be achieved on the Librispeech benchmark using pre-training on unlabeled data for speech processing. Interestingly, the method shows the capability of voice recognition with limited labeled data, outperforming earlier state-of-the-art approaches while using substantially less labeled data.

The wav2vec 2.0 framework demonstrates remarkable performance in a range of labeled data configurations, including ultra-low resource voice recognition using only 10 minutes of labeled data, according to experimental results. Additionally, the study assesses the framework's performance in labeled data sets with high resources, attaining a WER of 1.8/3.3 on the entire Librispeech benchmark. The study also provides a thorough examination of voice recognition error patterns and the effects of model size and unlabeled training data on the wav2vec 2.0 framework's performance. As audio input signals and EEG data share a waveform format, we hypothesize that using a similar method as outlined in \cite{baevski_wav2vec_2020} could produce encouraging results. In particular, we foresee the possibility of efficient EEG-to-text translation by training a generalized encoder model that can understand EEG features, similar to its function in executing speech-to-text tasks.

Some preliminary works on brain decoding were done from \cite{toneva_interpreting_2019} and \cite{wang_open_2021}, where embeddings derived from BERT play a central role in decoding human inner speech and thought processes. While \cite{toneva_interpreting_2019} successfully decoded inner speech using self-collected fMRI data, practical challenges associated with fMRI, such as non-real-time processing, continuous scanner availability, labour-intensive data collection, and substantial costs, underscore the need for alternative approaches.

In response to these challenges, \cite{wang_open_2021} pursued an alternative path, constructing a decoder analogous to \cite{toneva_interpreting_2019} but leveraging EEG data as the foundational dataset. This strategic shift aligns with the advantages offered by EEG, including cost efficiency, portability, and real-time data capture, making it a promising choice for decoding cognitive processes in real-world scenarios. 

Both referenced works have significant shortcomings. The analysis used in \cite{toneva_interpreting_2019} is based on fMRI data, which can be extremely costly to acquire and are not suitable for real-time processing. Similarly, instead of using raw EEG signals, \cite{wang_open_2021} uses word-level EEG characteristics obtained from the ZuCo dataset, which may restrict the model's application to real-time scenarios.

By using raw EEG signals, on the other hand, our method stands out and provides a model that may potentially decipher EEG data in real time. This crucial difference makes our approach a viable option for real-world scenarios needing instantaneous EEG-to-text translation.

\section{Preliminaries}

\subsection{Electroencephalography (EEG)}
An electroencephalogram (EEG) is a procedure designed to extract the electrical activity of the brain from the scalp. The recorded waveforms are presumed to reflect the dynamic activity of the brain's outer layer, known as the cerebral cortex. This region is considered to play a substantial role in shaping individual human thoughts, emotions, and behaviour. To monitor and record brain wave patterns, specialised sensors called electrodes are strategically affixed to the scalp at predefined locations. These electrodes are then connected to a computer system, allowing for the continuous monitoring and recording of electrical activity.

The positioning of electrodes follows the International 10/20 System, a standardised method for electrode placement on the head. The recordist, responsible for this procedure, carefully measures the head to ensure accurate electrode placement. Subsequently, signals captured by the electrodes are transmitted to the connected computer for recording, where digital EEG systems play a pivotal role. These systems transform the complex waveforms of EEG signals into a sequence of real-time numerical values, facilitating a more accessible and interpretable representation of brain activity. 

A critical parameter in this process is the sampling rate, denoting the frequency at which waveform data is sampled to convert it into a numerical format. Typically measured in hertz (Hz), the sampling rate, such as 512Hz, signifies the number of samples taken per second. This parameter is crucial for accurately capturing and representing the nuances of brain activity. The ability to observe millisecond-scale brain activity in real time stands out as one of the most remarkable advantages of EEG. This precision allows researchers to analyse the dynamics of brain function, providing valuable insights into cognitive processes.

\section{Methodology}
\label{methodology}

This section outlines the systematic approach used in this study for effectively training the brain-to-text decoder. It details the research framework and analytical techniques applied. Moreover, it offers an in-depth overview of the machine learning strategies and tools used, along with the rationale for their choice based on the outcomes of each integration. Additionally, it gives a brief description of the two datasets used and explains the logic behind their selection.

\subsection{Data}
During the development and enhancement of our Brain Decoder system, we implemented a strategy of training on two separate cerebral datasets. This approach was crucial for gaining a well-rounded understanding of neural activities across varied experimental setups. We initially used publicly available EEG datasets from ZuCo 1.0\cite{hollenstein_zuco_2018} and ZuCo 2.0\cite{hollenstein_zuco_2020} provided by the University of Zurich. These EEG datasets served as the fundamental base for our initial analyses and aided in creating a preliminary model of neural activity patterns, which became the basis for our subsequent research initiatives. Building on the initial insights from the EEG data, we decided to broaden our research by incorporating a second type of dataset, consisting of intracortical microelectrode arrays data. This additional dataset was sourced from the research conducted by Willett et al. \cite{Willett2023}.

Incorporating microelectrode array datasets allows for the capture of neural signals with significantly enhanced spatial and temporal precision compared to EEG data. This enhancement offers deeper understanding of cortical processes on a micro level.  This section offers a detailed explanation that supports our selection of these datasets. It includes a concise overview of each dataset, highlights their distinct features, and identifies the specific portions used as training data to improve the Brain Decoder's functionality.
\subsubsection{EEG}
\label{dataset_decoder}
This in-depth study utilizes the well-regarded and publicly available Zurich Cognitive Language Processing Corpus, namely the ZuCo 1.0 \cite{hollenstein_zuco_2018} and ZuCo 2.0 datasets \cite{hollenstein_zuco_2020}. These datasets are notable for their incorporation of both EEG data and eye-tracking information. The data is systematically collected from a substantial group of participants engaged in Normal Reading (NR) and Task-Specific Reading (TSR). The reading materials, which include detailed movie reviews and informative Wikipedia articles, have been carefully selected for analysis. A key feature of the ZuCo datasets is the precise temporal synchronization of EEG data with the text stimuli. This synchronization is achieved through detailed tracking of fixation points, accurately recorded via sophisticated eye-tracking devices. The datasets feature a broad array of EEG data intricately related to specific eye-tracking measures such as First Fixation Duration (FFD), which indicates the time spent on an initial fixation of a text segment.

Furthermore, the datasets provide information on total reading time (TRT), representing the entire duration of all fixations on a specific text, and gaze duration (GD), which measures the total time spent during the initial reading phase before any backward eye movements take place. The parameter known as single first fixation duration (SFD) indicates the time of an initial fixation on an object, whereas go-past time (GPT) is important for analyzing backward eye movements and patterns of re-reading in reading tasks.

The ZuCo corpus, known for its large-scale dataset, serves as a crucial resource for exploring cognitive language processing. By combining EEG and eye-tracking data, it allows researchers to study the complex interactions involved in reading behavior and brain activity. This fusion of data deepens our understanding of the biological and mental aspects of reading, thus fostering progress in brain decoding technology and cognitive neuroscience.

In this scholarly study, a detailed examination of raw EEG data on a sentence basis has been conducted. There are several strong reasons for this methodological choice. Primarily, sentence-level data aligns well with the structure of traditional speech-to-text datasets. This is a critical consideration, as explained in detail in the Methodology section of this dissertation. This section thoroughly clarifies how the design of the Brain Decoder system is fundamentally shaped by the principles of well-established speech-to-text and ASR frameworks, making sentence-level data particularly beneficial and pertinent.

Additionally, an in-depth examination of the dataset revealed notable word-level discrepancies. It became clear that while raw EEG data were captured for certain words, these recordings were occasionally absent for others. This irregularity varied across different participants. The original study's authors provide an explanation, suggesting that if a subject read a word faster than the EEG sampling interval, no data were recorded for that word during the related fixation period.

\subsubsection{Intra-cortical Microelectrode Arrays (IMA)}
Due to the discovery of inconsistencies in the ZuCo dataset, which cast doubt on its reliability and appropriateness for our research framework, an investigation was initiated to locate an alternative dataset that met the necessary criteria. The main criterion for our search was to find a dataset that mirrored the structure of ZuCo.

The alignment of the data setup permits smooth integration into our current analytical codebase, facilitating progress in our research without necessitating major changes or code restructuring. Our goal was to preserve the integrity and efficiency of our computations while minimizing any disruptions.

The dataset provided by \cite{Willett2023} was employed to enhance our investigation. This dataset involves a single participant reading natural text while intracortical arrays monitor their brain activity. Their study explores progress in neurolinguistics related to speech, particularly in developing assistive technologies for individuals with severe speech disabilities, such as ALS. A noteworthy aspect of the research is the development of a high-performance speech brain-computer interface (BCI) that deciphers neural signals related to speech production. This BCI demonstrates a word error rate of 9.1\% for a 50-word vocabulary and 23.8\% for a lexicon containing 125,000 words, with a communication speed of 62 words per minute, which is comparable to typical conversation rates.

This study highlights the importance of elements such as vocabulary size, electrode density, and training data for enhancing BCI performance, recommending an increase in microelectrodes and refinement of language models. Despite ongoing challenges related to system robustness and extended utilization, the research marks a notable advance in neurolinguistics, presenting opportunities to improve communication abilities for people with speech disabilities, ultimately enhancing their quality of life.

Even though the approach is invasive, it allowed for a more thorough assessment of our model, helping to ascertain if the technical issues found were responsible for ineffective model training.
\subsection{LLM substitution}
The foundation of our methodology was based on utilizing a stable transformer encoder, which was further improved by tweaking the activation function to facilitate text generation. In the beginning stages, our objective was to enhance the model's output quality by methodically substituting the LLM used at the end of the processing pipeline. We initially opted for BART as our primary LLM, acknowledging the constraints of the technological progress available during that period.

As research on LLMs progressed, a variety of innovative models have been incorporated into the machine learning framework. Meta AI's release of the Large Language Model Meta AI (LLaMa)\cite{meta2024introducing}, combined with OpenAI's later launches of GPT-3.5 and GPT-4.0\cite{openai2023gpt4}, marked a pivotal change in NLP. These trailblazing models showed remarkable abilities, outperforming previous standards and setting new benchmarks of achievement.

\begin{figure}[hbt!]
    \centering
    \includegraphics[width=\linewidth]{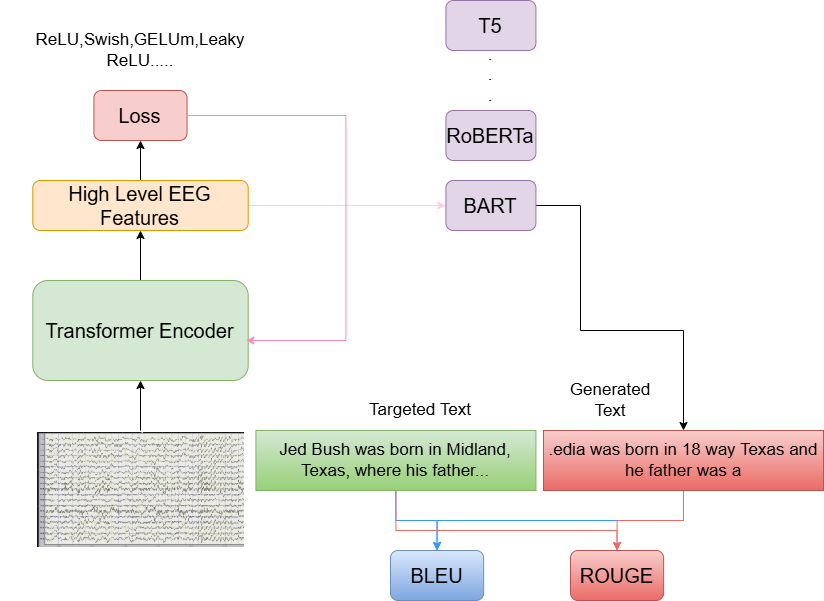}
    \caption{This figure shows the enhancement with the introduction of several state-of-the art LLMs. }
    \label{fig:brain_decoder_enhanced}
\end{figure}

Given our primary emphasis on text generation via LLMs, these sophisticated models provided a valuable chance to elevate the quality of text outputs surpassing BART's potential. We proposed that substituting BART with these cutting-edge models would enhance text generation. Accordingly, a systematic replacement of BART with these advanced models was undertaken, allowing for detailed documentation and analysis of our empirical findings during this developmental stage. Table \ref{tab:model_names_decoder} contains a complete list of all the models assessed within our pipeline.
\begin{table}[h]
    \centering
    \begin{tabular}{|c|}
        \hline
        \textbf{Model Name} \\ \hline
        EleutherAI-GPT-Neo-1.3B\cite{gpt-neo} \\ \hline
        Facebook-BlenderBot 400M\cite{roller2020recipes} \\ \hline
        Google BigBird Pegasus Large\cite{zaheer2021big} \\ \hline
        Microsoft Prophetnet Large Uncased\cite{yan2020prophetnet} \\ \hline
        RoBERTa\cite{zhuang-etal-2021-robustly} \\ \hline
        T5 \cite{2020t5} \\ \hline
    \end{tabular}
    \caption{List of Model Names  Tested as an enhancement to our pipeline.}
    \label{tab:model_names_decoder}
\end{table}

\subsection{CTC integration}
\label{sec:ctc_decoder}
In the beginning, our attempts to apply various LLMs did not yield substantial improvements in the pipeline's performance. This led to an in-depth analysis to identify potential improvements, ultimately revealing the benefits of integrating CTC loss\cite{graves2006connectionist}. CTC loss, widely used during the training of speech-to-text systems, effectively tackles the distinct attributes and format resemblances present in speech data, EEG, and IMA data. These modalities share a wave-based structure and are divided into time-steps based on sampling rates, which introduces challenges due to their variable input and target lengths. Such variations stem from differences in text sizes and variations in reading or speaking durations, which are affected by individual speeds and comprehension abilities.

The CTC loss function offers a comprehensive methodological approach for tackling the complex task of classifying unsegmented data. It effectively transforms sequences of varying lengths into coherent outputs. Unsegmented data, which lacks the distinct segmentation typically exemplified in auditory character depiction where the precise timing of the spoken element is known, presents a substantial challenge. The resolution of this challenge necessitates the adjustment of the sampling rate; however, variability remains, characterized by discrepancies arising when characters are articulated over extended time periods, influenced by complexity and speech speed. The process of segmenting brain data, which shares similar complexities, is notably arduous. Consequently, the incorporation of CTC loss into our framework was deemed optimal, as it successfully mitigates segmentation challenges. Within the domain of speech-to-text tasks, as documented, CTC loss has consistently demonstrated exceptional performance, often achieving notably superior results \cite{wang2017residual,graves2013speech}.

\begin{figure}[hbt!]
    \centering
    \includegraphics[width=\linewidth]{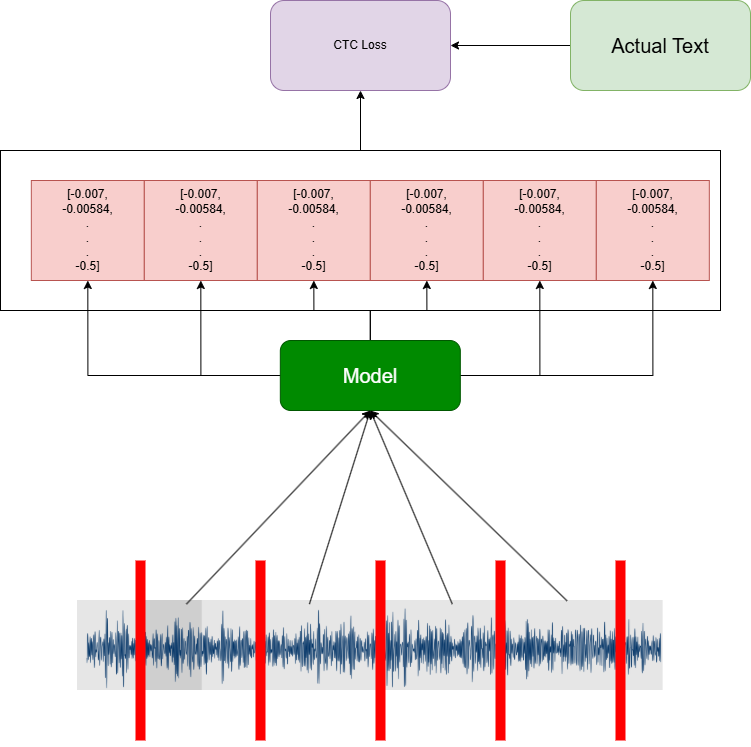}
    \caption{This figure illustrates how CTC loss was integrated to the pipeline with the hope of learning positional alignment of characters and brain data. For each time step the log probabilities of each character in the vocabulary were calculated and then the CTC loss was calculated between the predicted and actual sentence.}
    \label{fig:ctc_implementation_brain_decoder}
\end{figure}

This explanation clarifies why each time step is linked to a character, thereby simplifying the classification task and drastically decreasing the scope of classification in comparison to word-level classification, which relies on large dictionaries. The classification space is restricted to the 26 letters of the English alphabet, making it feasible to generate any possible word and thus broaden the vocabulary's range. Moreover, in spoken language, individual character sounds can shift, and combinations of characters can form unique phonetic sounds, such as the 'ch' sound in conversation. To manage these intricacies, phonemes were incorporated, significantly enhancing the efficacy of speech-to-text models. These insights led to the choice to tokenize our text data on a character or phonemic level for accurate classification in the timeframe of our data. To support this, we developed two custom tokenizers: one based on the English alphabet and the other using a phonetic vocabulary\cite{ieee9687921,hal-04584931}.

\subsection{Wav2Vec2 and Data2vec Implementation}
With the incorporation of CTC loss into our existing pipeline, we anticipated improved performance metrics. However, the anticipated boost in performance did not occur as expected. This lack of results is especially surprising given our use of cutting-edge architecture models. Specifically, we employed the Conformer model, which is well-documented, such as in the study by \cite{Gulati2020}, for delivering outstanding accuracy and efficiency in speech-to-text applications. The Conformer model is celebrated for its capability to handle sequential data inputs efficiently, adapting dynamically to the temporal variations found in speech signals.

The Conformer architecture combines CNNs and self-attention mechanisms, thus improving its ability to detect both local and global dependencies within acoustic signals. The model's design includes key features necessary for high performance in speech recognition systems, including scalability and resilience to perturbations and noise. Considering these qualities, it seemed logical to assume that integrating CTC loss into our pipeline would significantly enhance performance. However, the actual experimental outcomes differed from these expectations, leading to a deeper investigation of the factors preventing the anticipated performance improvements.

\begin{figure}
    \centering
    \includegraphics[width=0.5\linewidth]{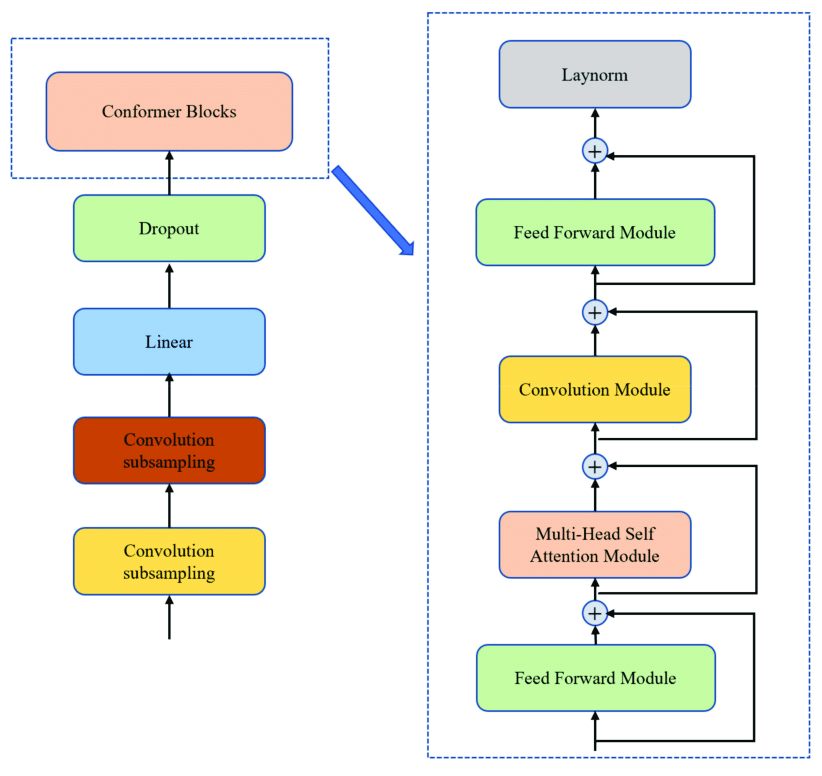}
    \caption{This figure shows the architecture of a Conformer Neural Network as proposed by Gulati et Al. \cite{Gulati2020} and how Convolution Layer can be integrrated with a Multi Headed Attention Layer.}
    \label{fig:conformer_architecture_brain_decoder}
\end{figure}

A comprehensive analysis of the training process for diverse models demonstrates that, frequently, data undergoes augmentation and feature extraction before it is fed into the conformer model. In the context of speech models, features are usually extracted by creating MEL spectrograms from audio files or by utilizing a pre-trained model on speech data to obtain features from specific audio samples. On the other hand, there is an absence of standardized processing methods for extracting features from EEG and IMA data. The current methods are not only complex but also demand substantial computational resources, requiring considerable processing power.

Wav2Vec2 \cite{Baevski2020} and Data2Vec \cite{Baevski2022} provide a training architecture focused on creating 'encoder models.' These models use self-supervised learning by structuring unlabeled data in a supervised fashion to extract general features relevant to different modalities. Data2Vec has been skillfully applied to various modalities, such as text, video, and images, to build strong encoders for each. Furthermore, due to the complexity and innovation in new machine learning models, these modalities can be seamlessly combined to create powerful multimodal models that can process a diverse range of data types effectively.

\begin{figure}[hbt!]
    \centering
    \includegraphics[width=\linewidth]{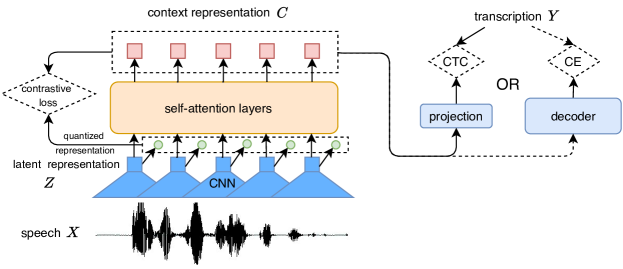}
    \caption{This Figure shows the proposed architecture for a Wave2Vec2 model training regime as proposed by \cite{Baevski2020}}
    \label{fig:wav2vec_architecture_brain_decoder}
\end{figure}
\begin{figure}[hbt!]
    \centering
    \includegraphics[width=\linewidth]{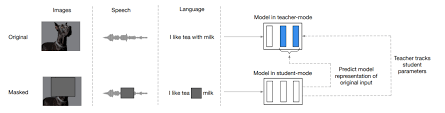}
    \caption{This figure illustrates the proposed architecture for training effectively a Data2Vec model as proposed by \cite{Baevski2022}}
    \label{fig:enter-label}
\end{figure}

In our efforts to create a model that replicates brain-like data, we identified a notable gap: existing models, as far as we know, have yet to achieve this goal. This gap led us to investigate using Data2Vec to develop a brain modality that is genuinely reflective. Our experimental model primarily utilizes the well-known Conformer architecture, extensively illustrated in Figure \ref{fig:conformer_architecture_brain_decoder}. Conformer blocks offer a distinct edge due to their hybrid structure, adeptly merging the feature extraction prowess of CNNs with the comprehensive attention mechanisms of transformer architectures. This is achieved through a deliberate arrangement of self-multiheaded attention layers within each conformer block. Our model's design incorporates multiple Conformer blocks in a hierarchical manner, with each layer adding to the model's complexity. The adjustable number of blocks allowed us to systematically enhance model intricacy and detect subtler patterns in our data. Following the conformer block arrangement, the architecture transitions into a fully connected layer, ending with a projection layer designed to output character probabilities within our specified classification framework. The use of a Log Softmax activation function here is particularly significant. 

In contrast to the conventional Softmax approach, Log Softmax presents distinct advantages, particularly within the realm of machine learning applications such as neural networks. The excellence of Log Softmax is evidenced by its enhancement of numerical stability, a common issue arising from overflow and underflow related to large score values (logits) in neural computations. By applying the log-sum-exp trick, Log Softmax adeptly mitigates these challenges, thereby transforming the Softmax and logarithm computations into a more streamlined process that augments computational efficiency. This heightened efficiency contributes to accelerated convergence during training by improving the penalization of errors in inaccurate predictions, thereby fostering a more precise gradient descent path. Moreover, the amalgamation of Log Softmax with loss functions, particularly cross-entropy, underscores its inherent compatibility. As these loss functions require logarithmic probabilities, Log Softmax optimizes the training process by not only ensuring stable gradient computation, which is vital for robust back-propagation, but also by adeptly accommodating extreme data values. These characteristics yield a balanced probability output, ultimately establishing Log Softmax as an astute choice in various deep learning contexts.

\subsection{Bendr And EEG-Conformer Integration}
To enhance and refine the existing processing pipeline, significant changes were made by integrating two versatile EEG encoding models: Bendr\cite{kostas_bendr_2021} and EEG-Conformer\cite{9991178}. Originally developed for creating generalized EEG features, these encoders serve specific roles: Bendr is mainly used for EEG feature classification tasks, while EEG-Conformer aims to capture both local and global features within a single EEG classification framework.

It is important to note that neither encoding model was initially designed for the purpose of producing text directly from EEG data. However, it was hypothesized that with careful adjustments, these models could be refined to meet the unique demands of our specific task. To achieve this adjustment, a change to the existing architecture was made. This change involved the integration of an extra pair of layers at the final phase of the pipeline, specifically a fully connected layer followed by a projection layer. The main goal of this structural change was to accurately compute and generate a probability distribution over potential character outputs, thereby enhancing the model's ability to generate text from EEG data.

\section{Results}
\label{results}
In evaluating the effectiveness of our models, we found it fitting to use the BLEU score and ROUGE score as our primary metrics. Due to the consistent nature of the task, we considered it prudent to continue employing these metrics for two key reasons. First, as outlined earlier in this thesis, the BLEU and ROUGE scores are recognized as among the most respected benchmarks in the text generation evaluation field. Their reputable status in the discipline supports our decision.

These metrics were crucial in assessing the foundational transformer Encoder. As such, their ongoing application in evaluating our models supports a coherent and smooth comparison of results across different iterations. Keeping this consistency is vital for correctly interpreting performance advancements and the effectiveness of our new methods within the boundaries of established benchmarks.

\begin{table}[h]
  \centering
  \resizebox{\linewidth}{!}{%
    \begin{tabular}{|l|l|l|l|}
      \hline
      \textbf{Scenario} & \textbf{Technique}        & \textbf{EEG (BLEU, ROUGE)} & \textbf{IMA (BLEU, ROUGE)} \\ \hline
      \multirow{2}{*}{CTC}  
                       & CTC+Phoneme  & (0.02, 0.0)  & (0.01, 0.0) \\ \cline{2-4} 
                       & CTC+Character & (0.09, 0.0)  & (0.07, 0.001 ) \\ \hline
      \multirow{4}{*}{Generic Algorithms}  
                       & Data2Vec+Phoneme  & (0.02, 0.0005)  & (0.008, 0.0005)  \\ \cline{2-4}  
                       & Data2Vec+Character  & (0.1, 0.02)  & (0.05, 0.02)  \\ \cline{2-4} 
                       & Wav2Vec2+Phoneme  & (0.02, 0.0002)  & (0.008, 0.0005)  \\ \cline{2-4} 
                       & Wav2Vec2+Character  & (0.1, 0.02)  & (0.05, 0.02)  \\ \hline  
      \multirow{4}{*}{Brain Encoders}  
                       & Bendr+Phoneme  & (0.03, 0.0002)  & (0.02, 0.0)  \\ \cline{2-4}  
                       & Bendr+Character  & (\textbf{0.13}, 0.0)  & (0.04, 0.0)  \\ \cline{2-4}  
                       & EEG-Conformer+Phoneme  & (0.02, 0.0004)  & (0.009, 0.0006)  \\ \cline{2-4} 
                       & EEG-Conformer+Character  & (0.1, \textbf{0.02})  & (0.05*, 0.0)  \\ \hline 
    \end{tabular}%
  }%
  \caption{Performance comparison of different techniques and scenarios using EEG and IMA metrics (BLEU, ROUGE).}
  \label{tab:results_brain_decoder}
\end{table}

Our empirical results clearly indicate that simply swapping the LLM in the current pipeline did not lead to a noticeable enhancement in outcomes. This aligns with prior research, which suggests that LLMs are mainly used as correction tools rather than for transformation when converting different modalities into text\cite{brown2020language,lei2024contextualization,zhao2024generative}. For example, in the context of transcribing spoken words to written text, words like "red" and its homophone "read" (past tense) remain phonetically identical.

Given these circumstances, determining the intended meaning requires leveraging the sentence structure and contextual cues that a decoder cannot access without a properly pre-trained LLM. As a result, while the LLM plays a crucial role as a key component in the system, it is crucial that the text generation facilitated by the decoder reaches a level of sophistication necessary for accurately extracting and generating coherent and contextually relevant sentences. Therefore, incorporating an LLM is essential to ensure translations are both syntactically correct and semantically meaningful. Additionally, when checking our pipeline for errors or issues, it became clear that teacher forcing impacted our pipeline, invalidating our results. Therefore, considering these factors, we decided to discard our results, as they were invalid and outside the scope of our study.

Due to the reliance of large language models (LLMs) on the decoder element, we initiated the incorporation of CTC loss into our process. This step was driven by the goal of creating a more efficient decoder model before launching the LLM. As outlined in our methodology Section \ref{sec:ctc_decoder}, CTC loss is a widely acknowledged strategy in speech-to-text applications, which face similar data format issues as our task, thereby validating our choice of this specific loss function.

Additionally, the speech-to-text domain often grapples with the issue of variable-length input recordings, a problem similar to the challenges we encounter. The efficiency of CTC is shown through its successful handling of such variability. However, our own tests using solely the CTC loss did not yield the anticipated enhancement in our decoder's performance, as illustrated in Table \ref{tab:results_brain_decoder}. We suspect that the stagnation is attributed to the lack of an Encoder system, specifically a more targeted Brain Encoder.

It is crucial to position this research within the existing body of literature, particularly focusing on the encoder-decoder framework, as it may offer deeper insights into our hypothesis. The encoder-decoder architecture was chosen because it has repeatedly demonstrated exceptional performance in various sequence-to-sequence tasks, as previously evidenced in the literature \cite{bahdanau2015neural,cho2014properties,Sutskever2014}. Given that our task involves generating a text sequence from a brain sequence, this architecture is ideally suited. This foundational gap highlights the potential necessity of incorporating an encoder to achieve improved results. Consequently, although CTC loss presents certain advantages, its application in isolation is inadequate, thereby necessitating the investigation of an augmented architecture to obtain superior outcomes.

To evaluate the hypothesis presented earlier, a range of methods was used to train various encoders, which were then assessed within our system. Initially, the encoder model was trained using two mainstream approaches: Data2Vec and Wave2Vec2. These methods surpass the existing state-of-the-art across different modalities, including videos, text, and audio, providing distinct insights into crafting modality-agnostic encoders. However, we couldn't generate results for these two modalities due to a specific issue. The CTC loss recorded at the final step was negative. A negative CTC indicates that some probabilities at each time step, computed using log softmax, are positive. Probabilities from log softmax should remain negative, as it reflects the negative log-likelihood. We proposed that this issue arises due to the encoder failing to learn accurate representations, thus yielding suboptimal features for the decoder.

To tackle the challenges posed by encoder obstacles, we utilized two pre-trained models: BENDR \cite{kostas_bendr_2021} and EEG-Conformer \cite{9991178}. Initially created as versatile EEG encoders for multiple datasets, these models required only slight modifications for classification purposes. Likewise, we implemented minimal changes to fine-tune these models on our datasets and then evaluated their effectiveness as encoders within our processing framework. As shown in Table \ref{tab:results_brain_decoder}, employing both encoders provided results; however, these were inadequate for successful brain-to-text decoding.

During our analysis, we discovered an inconsistency previously ignored in our dataset: some cases in Table \ref{tab:results_brain_decoder} lacked data in the results. A detailed review of the execution logs indicated that during the model's training, the loss metric stayed persistently high. Normally, loss metrics should vary between 0.1 and 1.0, reflecting expected model performance. However, we noticed loss values exceeded 2.0, with some training runs showing numbers reaching 30 or more. These high loss values led to 'gradient explosion,' where the model's gradients diverged, resulting in infinite loss calculations. This issue disrupted the training, preventing the model from learning efficiently from the data. This finding is important as it points to possible issues in the training process that could hinder model convergence, offering insights into how we might improve our approach in future experiments.

\section{Discussion and Conclusion}
\label{conclusion}
In this investigation, we examined the feasibility of modifying existing encoding models to directly transform EEG data into textual representation. This examination prompted us to alter the current model architecture by incorporating an additional fully connected layer alongside a projection layer. The evaluation of these modifications was conducted utilizing well-established metrics, including BLEU and ROUGE scores, which are highly esteemed in the domain of text generation.

The outcomes, as presented in Table \ref{tab:results_brain_decoder}, indicate that a mere modification of the language models (LLMs) incorporated into the EEG-to-text pipeline does not substantially improve transcription accuracy. This corroborates prior research suggesting that LLMs are conventionally more efficacious for correction than for tasks involving modality transformation.

A crucial phase in our research involved assessing the efficacy of Connectionist Temporal Classification (CTC) loss within our framework, given its prevalent application in the speech-to-text domain. Despite CTC's potential in managing variable-length inputs, it did not result in a significant enhancement of our decoder's performance when applied independently. Consequently, this led to an investigation of Brain Encoders, including BENDR and EEG-Conformer; however, these models did not achieve the desired level of proficiency in brain-to-text conversion.

During our experimental procedures, we encountered several significant challenges. Specifically, we consistently observed elevated loss metrics throughout the model training phase, which induced gradient explosions and inhibited effective learning. Moreover, the lack of Encoder components capable of acquiring precise representations substantially contributed to the stagnation in performance.

In conclusion, although BLEU and ROUGE metrics facilitated a uniform assessment of our models, it became apparent that the current encoder architecture lacks the sophistication necessary for accurately translating EEG data into coherent textual sequences. Our analysis indicates that progressing brain-to-text systems necessitates the development of more innovative encoder-decoder frameworks capable of accommodating the inherent variability and complexity of EEG signals. Future research should address the identified challenges by prioritizing the development of robust encoder systems, thereby establishing a strong foundation for the efficient generation of text from brain data.
\label{conclusion}

\bibliographystyle{ACM-Reference-Format}
\bibliography{biblio}
\end{document}